\documentclass[12 pt]{article}

\usepackage{graphicx}

\usepackage{color}
\usepackage{latexsym}
\usepackage{amssymb}
\usepackage{amsmath}
\usepackage{mdframed}

\usepackage{hyperref, url}
\usepackage[show]{ed}
\usepackage{subfigure}
\usepackage[ntheorem]{empheq}
\usepackage{amsthm, amssymb, amsfonts, latexsym}
\makeatletter
\DeclareRobustCommand{\cev}[1]{%
  \mathpalette\do@cev{#1}%
}
\newcommand{\do@cev}[2]{%
  \fix@cev{#1}{+}%
  \reflectbox{$\m@th#1\vec{\reflectbox{$\fix@cev{#1}{-}\m@th#1#2\fix@cev{#1}{+}$}}$}%
  \fix@cev{#1}{-}%
}
\newcommand{\fix@cev}[2]{%
  \ifx#1\displaystyle
    \mkern#20mu
  \else
    \ifx#1\textstyle
      \mkern#20mu
    \else
      \ifx#1\scriptstyle
        \mkern#26mu
      \else
        \mkern#26mu
      \fi
    \fi
  \fi
}

\makeatother

\newcommand{\fig}[1]{\epsfbox}

\newtheorem{Theorem}{Theorem}[]
\newtheorem{Lemma}{Lemma}[section]

\newtheorem{Corollary}{Corollary}[section]

\begin{document}

\begin{center} 
This is the article to appear in the Proceedings of the Royal Society of London, Series A, DOI: \url{10.1098/rspa.2020.0563}
\end{center}

\title{
Stability and Memory-loss go Hand-in-Hand: Three Results in Dynamics \& Computation}

\author{G. Manjunath \\
{\small \it Department of Mathematics \& Applied Mathematics, University of Pretoria, Pretoria 0028} \\ {\small Email: manjunath.gandhi@up.ac.za } \\}
\date{}

\maketitle

\begin{abstract}The search for universal laws that help establish a relationship between dynamics and computation is driven by recent expansionist initiatives in biologically inspired computing. A general setting to understand both such dynamics and computation is a driven dynamical system that responds to a temporal input. Surprisingly, we find memory-loss a
feature of driven systems to forget their internal states helps provide unambiguous answers to the following fundamental stability questions that have been unanswered for decades: what is necessary and sufficient so that slightly different inputs still lead to mostly similar responses? How does changing the driven system's parameters affect stability? What is the mathematical definition of the edge-of-criticality?  
We anticipate our results to be timely in understanding and designing biologically inspired computers that are entering an era of dedicated hardware implementations for neuromorphic computing and state-of-the-art reservoir computing applications. 

\end{abstract}


\section{Introduction}
 
The ability of nearly all living systems to process and act on external stimuli indicates life's capabilities of computing (e.g., \cite{bray2009wetware}). In the human-made world, we are in the era of 
having millions of artificial neurons on silicon chips, and building biological computational devices using engineered biological units like yeast cells  (e.g., \cite{regot2011distributed}). The intriguing question is what exactly is required by these computing paradigms, be it from the simplest to the most complex to respond distinctly and yet have input-related stability, i.e., slightly different versions of the input should lead only to slightly different responses. 
The criterion of criticality employed in computation (e.g., \cite{kauffman1993origins,langton1990computation,kitzbichler2009broadband}), often deemed a governing principle of certain brain dynamics seems to be one of the successful paradigms in forgetting internal states while responding to the external stimuli or input.   However, the question remains as to whether such computation always exhibits input-related stability? Moreover, next, is there any other computing paradigm that provides input-related stability? This article aims to provide conclusive answers to these questions.

In analogy with how a substance like water that raises its temperature by excitation due to external heat and also can only dissipate heat if its temperature is above a threshold (see \cite{SI} for a thought experiment), a driven dynamical system can exhibit two competing dynamical responses, one that allows the temporal input to excite the system in the sense the system states are made to scatter apart at certain times and the other to quench the input by bringing states together at some other times.  Mathematically speaking, such contrasting dynamics requires nonlinearity, and the quenching would result in the dynamics be bounded.   Systems that are nonlinear and that have bounded dynamics are abundantly found in many models. They 
include ecosystems responding to external signals ranging from predator-prey models to tumor growth (e.g., \cite{murray2007mathematical}), many classes of artificial systems ranging from open-loop control systems (e.g., \cite{terrell2009stability}) to neural network models used in machine learning and artificial intelligence encompassing the field of reservoir computing  (e.g., \cite{lukovsevivcius2009reservoir}), in all of which the external input could also be by design.

Memory-loss in such driven systems appears when the net effect of the bringing states together and scattering the states of the system results in the diameter of the state asymptotically decreasing to zero so that all trajectories tend to coalesce into a single trajectory (see \cite{SI} for examples).   In that spirit, we call a system to have a memory-loss if a single trajectory emerges as a proxy of the input, or in other words, echoes the temporal input in the state space. Memory-loss is a property of both the temporal input and the system, a fact unfairly neglected mostly in the literature. 
The idea of relating memory-loss to the presence of a single proxy corresponding to an input is illustrated for both continuous-time (see \cite[Section 1]{SI}) and discrete-time dynamical systems as the echo-state property with respect to an input (representative work is in \cite{Manju_ESP}).  Much of the literature considers the notion of memory loss to all possible input sequences. Although these ideas date back to the works in \cite{wiener1966nonlinear,volterra1959theory},
they were formalised as different notions such as the  fading memory property in  \cite{boyd1985fading}, the input forgetting property in \cite{chua1976qualitative}, the echo state property, the state contraction, the state forgetting property in \cite{jaeger2001echo}. A rigorous treatment of several of these notions is available in \cite{grigoryeva2018echo,grigoryeva2019differentiable}. The idea of exploiting memory-loss for applications dates back to work in  \cite{kocarev1996generalized}, however, realizing it for computation and information processing was conceived independently by two research groups -- Jaeger et al. in  \cite{jaeger2001echo,jaeger2004harnessing} as the Echo State Property (ESP) and Maass et al. in \cite{maass2002real} as the Liquid State Machine. Collectively the methodologies are referred to as reservoir computing (RC), and some class of these systems can approximate any input-output function to arbitrary precision \cite{grigoryeva2018echo}. Today, RC has touched upon the era of hardware implementations (e.g., \cite{appeltant2011information, kudithipudi2016design, vandoorne2014experimental}) with photonic chips, memristors, spintronic nanodevices with even applications such as in prosthetic control and epileptic seizure detection besides classical application of prediction, filtering and information processing of temporal data (e.g., \cite{lukovsevivcius2009reservoir} and references therein).

A fundamental question in computing or modeling by driven dynamical systems is to find if temporal inputs are mapped \emph{continuously} onto the state space of the dynamical system -- input-related stability -- failing which,  certain temporal patterns in the input could drive the system into an unstable regime where it would behave wildly in the sense that small changes in the input render a drastically different system response. Input-related stability being a function of both the system and the input, in practice, a problem is either to examine if an input given a system, or if a design parameter of the system given an input, renders input-related stability. Another problem would be to understand the change in the design parameter -- parameter-related stability for a given input -- failing which, change in the parameter could give a drastically different response. For a representation in the state space to capture the information content of the input, better separability of inputs in the state space is needed (see example in \cite[Section 1]{SI}).  In examples such as in RC, one would be interested in maximizing the freedom of the driven system, i.e., maximize the separability of the inputs to distinguish close-by-variants of an input in the state space of the system, but yet not tipping the system into what authors following \cite{langton1990computation} have called it the ``chaotic" behavior.  
While RC is used in a task of classifying inputs, such tipping reflects on a trade-off between input separability and input-related stability. This is also the hallmark of criticality, and such close to criticality operation has been found to be self-organized, like in biological neural systems (e.g., \cite{beggs2008criticality}), or can be tuned into, in artificial systems like RC. 
The criticality hypothesis drives the need to operate these systems under high-efficiency and thus to choose the design parameters more optimally and yet avoid instabilities. The challenge of doing so is overwhelming since there is no clear understanding of how the entire temporal input is mapped onto a different space, which is usually higher-dimensional.

A design parameter (not necessarily a scalar and includes the case of a vector or a matrix) $\alpha$, an input value $u$, a state-variable $x$ in a compact state space $X$ and a continuous function $g(\alpha,u,x)$ that takes values in $X$ would constitute a parametric driven system (we denote this system by $g$ throughout, and other entities silently understood).   The dynamics of a parametric driven system $g$ is rendered through a temporal input $\bar{u} = (\ldots ,u_{-1},u_0,u_1,\ldots)$ via the equation $x_{n+1} = g(\alpha,u_n,x_n)$. Given 
an $\bar{u}$ and $\alpha$, a sequence $\bar{x} =  (\ldots ,x_{-1},x_0,x_1,\ldots)$ contained in $X$  that satisfies the equation $x_{n+1} = g(\alpha,u_n,x_n)$ \emph{for all } 
integers $n$ is an image of $\bar{u}$ in the space $X$ (and also called a solution of $g$ for a given $\bar{u}$). Having fixed  $\alpha$, suppose we collect the $n^{\mbox{th}}$ component of all images $\bar{x}$ of an $\bar{u}$, and denote it by $X_n(\alpha,\bar{u})$, then we have a collection of sets  
$\{X_n(\alpha,\bar{u})\}$ which we call it the representation of $\bar{u}$.  
Following the intuitive idea of memory-loss as in \cite{Manju_ESP}, we say that the driven system $g$ has echo-state property (ESP) with respect to ( w.r.t.\ ) an input  $\bar{u}$ at $\alpha$ if it has exactly one solution or one image $\bar{x}$ in the space $X$ or (equivalently) if for each $n$, $X_n(\alpha,\bar{u})$ is single-valued, i.e., it is a singleton subset of $X$, i.e., there is no possibility of two or more reachable states at any given point in time $n$ if the entire past values of the input have been fed into the driven system.

\begin{center}
\begin{figure}[h]
\centering
\includegraphics[width=9cm]{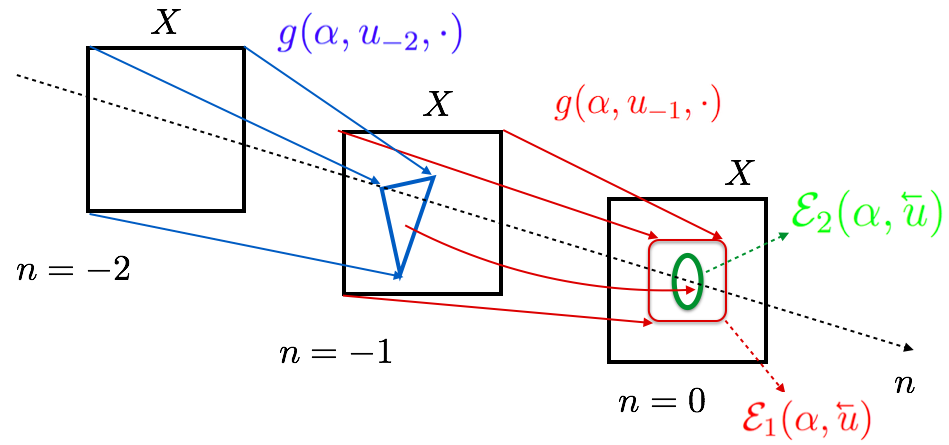}
\caption{Schematic figure to explain the sets 
$\mathcal{E}_n(\alpha,\cev{u})$: When $X$ is a  square (boundary in black), $\mathcal{E}_1(\alpha,\cev{u}) = g(\alpha,u_{-1},X)$ is the  square-like object (boundary in red), and when $g(\alpha,u_{-2},X)$ is the triangle (boundary in blue) then $\mathcal{E}_2(\alpha,\cev{u}) = g(\alpha,u_{-1},g(\alpha,u_{-2},X)$ is the  oval (boundary in green).} 
\label{Fig_intersection2}
\end{figure}
\end{center}

We observe that $X_0(\alpha,\bar{u})$ can be determined only by the left-infinite segment of the input $\cev{u} := (\ldots, u_{-2},u_{-1})$. Hence, we define $\mathcal{E}(\alpha, \cev{u}): = X_0(\alpha,\bar{u})$ as the encoding of $(\alpha,\cev{u})$.  The set $\mathcal{E}(\alpha, \cev{u})$ turns out (see Eq. \eqref{eqn_Xn}) to be the countable intersection of the sets $\mathcal{E}_1(\alpha, \cev{u}) :=  g(\alpha, u_{-1},X)$, $\mathcal{E}_2(\alpha, \cev{u}) :=  g(\alpha, u_{-1},g(\alpha, u_{-2},X))$ and so on (see 
Fig.~\ref{Fig_intersection2} for a schematic description). Further, $\mathcal{E}(\alpha, \cev{u})$ happens to be also the limit of the sequence of sets $\mathcal{E}_1(\alpha, \cev{u}), \mathcal{E}_2(\alpha, \cev{u}), \ldots$ (see Eq.\eqref{eqn_Xn2}). 
Given the input segment $(u_{-n},\cdots, u_{-2},u_{-1})$ of an $\cev{u}$ and a subset of $Y_0\subset X$, consider the set $Y_n$ recursively obtained by $Y_1 = g(\alpha, u_{-n},Y_0)$, $Y_2= g(\alpha, u_{-n+1},Y_1)$, and so on. From the viewpoint of computer simulations, if the cardinality of $Y_0$ is sufficiently large,  the set $Y_n$ is an approximation of  $\mathcal{E}_n(\alpha, \cev{u})$, and due to the aforementioned limit, $Y_n$ is an approximation of $\mathcal{E}(\alpha, \cev{u})$ as well.

Our approach is to understand the influence of $\cev{u}$ and $\alpha$ separately on the encoding  $\mathcal{E}(\alpha, \cev{u})$. To do this, we vary $\cev{u}$ keeping $\alpha$ constant, i.e., study the set-valued function $\cev{u} \mapsto \mathcal{E}(\alpha, \cev{u})$, which we call it the \emph{input-encoding map (given $\alpha$)}. Next, we vary $\alpha$ keeping $\cev{u}$ a constant, i.e., study $\alpha \mapsto \mathcal{E}(\alpha, \cev{u})$, which we call it the \emph{parameter-encoding map (given $\cev{u}$)}. While we study the entire representations in these two cases, i.e., when we study $\bar{u} \mapsto \{X_n(\alpha,\bar{u})\}$ for a given $\alpha$ and $\alpha \mapsto \{X_n(\alpha,\bar{u})\}$ for a given $\bar{u}$, we call these functions the \emph{input-representation} and \emph{parameter-representation} maps respectively.

Both these encoding maps are set-valued functions, and in general, a set-valued map $S(x)$ is continuous at $x$ when it is both lower semicontinuous (l.s.c) and upper semicontinuous (u.s.c) that respectively precludes explosions and implosions at $x$ (see \cite[Fig. S3]{SI} for a schematic picture and definition). We remark that the continuity of a set-valued map $S(x)$ can be equivalently defined by employing 
the Hausdorff metric (see Section~\ref{Sec_MD}).

We make two natural assumptions on driven systems that ensure the existence of a subdomain where states do not ``collapse" to a single point but can contract the state space volume. We say a parametric driven system $g$ to be an \emph{open-parametric} driven system if $X$ is connected and the mapping $g(u,\alpha,\cdot)$ does not map any set with uncountably many points contained in $X$ to a single element in $X$. This further ensures that any non-zero state space volume does not ``suddenly collapse". If $g(\alpha, u, .)$ is invertible for all parameter-values $\alpha$ and input values $u$ under consideration, then $g$ is obviously an open-parametric driven system.  Also, one can easily show that an open-parametric driven system $g$ has the ESP w.r.t.\ $\bar{u}$ (or to $\cev{u}$) {\bf if and only if} $\mathcal{E}(\alpha, \cev{u})$ is a singleton subset of $X$ (see Section~\ref{Sec_MD} for details). Further, if a driven dynamical system has a ``contracting subdomain" in its state space for each $\alpha$, we characterize it by the terminology of contractibility.  Precisely, given an open-parametric driven system $g$, and an $\alpha$, if there exists an input $\bar{v}$ such that $g$ has the ESP w.r.t.\ $\bar{v}$ at $\alpha$, then we say that $g$ is \emph{contractible} at $\alpha$.  Contractibility does not mean a contraction mapping but is a condition that ensures the existence of an input $\bar{v}$ so that for every parameter $\alpha$,  $\bar{v}$ can push the dynamics into a subset of $X$ so that we have a net contraction in the sense $g$ has the ESP w.r.t.\ $\bar{v}$.
The example of a recurrent neural network (see Section~\ref{Sec_Results}) is an example of a driven system that satisfies these two hypotheses for our numerical simulations.

The paper is organized as follows. In Section~\ref{Sec_Results}, we state three results concerning stability and criticality of driven systems with less formalism and illustrate them using numerical results obtained from recurrent neural networks. In Section~\ref{Sec_MD}, we restate our results formally and prove them -- the mathematically inclined reader may choose to peruse Section~\ref{Sec_MD} prior to reading Section~\ref{Sec_Results}. In Section~\ref{Sec_Discussion}, we summarise the results and mention their broader implications. The supplementary material \cite{SI} available online contains a discussion on the intuitive meaning of the memory loss property, a slightly more elaborate version of Section~\ref{Sec_MD}  and additional numerical results.

\section{Three Stability Results} \label{Sec_Results}

Mathematically, the continuity of the input-encoding map at $\cev{u}$ means input-related stability at $\cev{u}$ and the continuity of the parameter-encoding map at $\alpha$ means parameter-related stability at $\alpha$. The mathematical results are in Section~\ref{Sec_MD}, and here we state the results with less formalism and illustrate them with numerical results. 
Additional details used in obtaining the numerical results presented here are available in \cite[Section 6]{SI}.

To illustrate our results, we consider the example of a recurrent neural network (RNN). RNNs are employed for processing temporal data and are inspired by how the brain maps a stimulus onto its enormously higher-dimensional space by affecting the states of a vast number of neurons.  An advantage observed in projecting data onto a higher-dimensional space is that it exaggerates certain data features to render better separability of inputs.  In the RC framework employing RNNs, the temporal input signal is mapped on to a higher dimensional space (reservoir) without training. And then, a simple regression technique is used to obtain
linear observables of the higher-dimensional entity that aims at an approximation of the desired function of the temporal signal (e.g., \cite{jaeger2001echo}).  We consider RNNs that are driven systems of the form $g(\alpha,u,x) = \overline{\tanh}(Au + \alpha Bx)$, where $\overline{\tanh}(*)$ is (the nonlinear activation) $\tanh$ performed component-wise on $*$, $\alpha$, a real-valued parameter would correspond to the scaling of the reservoir (invertible) matrix $B$ of dimension $N$ and $A$ is a matrix with input connections and $X=[-1,1]^N$ (the cartesian product of $N$ copies of $[-1,1]$). To maintain uniformity across numerical experiments, we set $B$ to have a unit spectral radius. The driven dynamics through the equation $x_{n+1} =g_\alpha(u_n,x_n) = \sigma(A u_n+ \alpha B x_n)$ is interpreted to be representing the evolution of an ensemble of neurons (coordinates of $X$).  The function $\overline{\tanh}$ is an example of a saturation or an activation function $\sigma$ that is nonlinear enough to``squash" the system state to a smaller subspace.

Before, we state our results, we observe that whenever $\alpha \not = 0$ and $B$ is invertible, $g(\alpha,u,\cdot)$ is always invertible 
since $\overline{\tanh}$ is invertible on $(-1,1)^N$, and hence the RNN is an open-parametric driven system when $\alpha =0$ is not in the parameter space $\Lambda$. 
Also, the RNNs that we consider are always contractible except in the case of the input weight matrix $A$ having a row with all zeroes. It is a result in \cite[(ii) of Theorem 4.1]{Manju_ESP} that for a RNN with an input-weight matrix $A$ having non-zero rows and any norm of the reservoir matrix $B$, there always exists an input $\cev{v}$ so that it has the ESP w.r.t.\ $\cev{v}$, and hence such systems are contractible. The idea behind contractibility is that if the input $\cev{v}$ is such that if the entity $(Av_n + \alpha Bx_n)$ is thrown into the saturation region of the activation function $\sigma$ sufficiently often, one can always witness ESP w.r.t.\ $\cev{v}$. Given any $\cev{u}$, an $\cev{v}$ is readily obtained by scaling $\cev{u}$ to have a sufficiently large amplitude.

For an open-parametric dynamical system $g$ we present the following three results {\bf (R1)}, {\bf (R2)} and {\bf (R3)} that are proved as Theorem~\ref{Thm_Main1}, Theorem~\ref{Thm_Main2} and Theorem~\ref{Thm_Main2b} respectively in Section~\ref{Sec_MD}: 
{\bf (R1)}. If $g$ is contractible at $\alpha$, then the input-encoding map. i.e., $\mathcal{E}(\alpha, \cdot)$ and the input-representation map $\{X_n(\alpha,\cdot)\}$ are both continuous at $\cev{u}$ if and only if $g$ has the ESP w.r.t.\ $\cev{u}$.
{\bf (R2)}.  If $g$ has the ESP w.r.t.\ $\cev{u}$ then 
both the parameter-encoding map. i.e., $\mathcal{E}(\cdot, \cev{u})$ and the parameter-representation map $\{X_n(\cdot,\bar{u})\}$ are both continuous at $\cev{u}$. 
{\bf (R3)}. The parameter-encoding map $\mathcal{E}(\cdot,\cev{u})$ is continuous at $\beta$ if and only if $\{\mathcal{E}_n(\cdot,\cev{u})\}$ is equicontinuous at $\beta$.

The result {\bf (R1)} states that ESP is fundamental to avoid instabilities (discontinuous changes) due to small changes in the input in a wide-class of driven systems.   The contractibility condition produces dissipative dynamics that resets the dynamics in every coordinate of the state space (a neuron in the case of a RNN), thereby inducing memory-loss. This is illustrated intuitively in a thought experiment in \cite{SI}. Further to {\bf (R1)}, we observe in Corollary~\ref{Corollary_Continuity} that the ESP always implies input-related stability in systems that are not necessarily contractible. However, without the contractibility hypothesis, {\bf (R1)} would not hold, for example, the driven system $g(\alpha, u,x)=x$ has $\mathcal{E}(\alpha, \cev{u})=X$ for all $\cev{u}$ and hence
the input-encoding map is continuous.

\begin{center}
\begin{figure}[h]
\centering
\includegraphics[width=15cm]{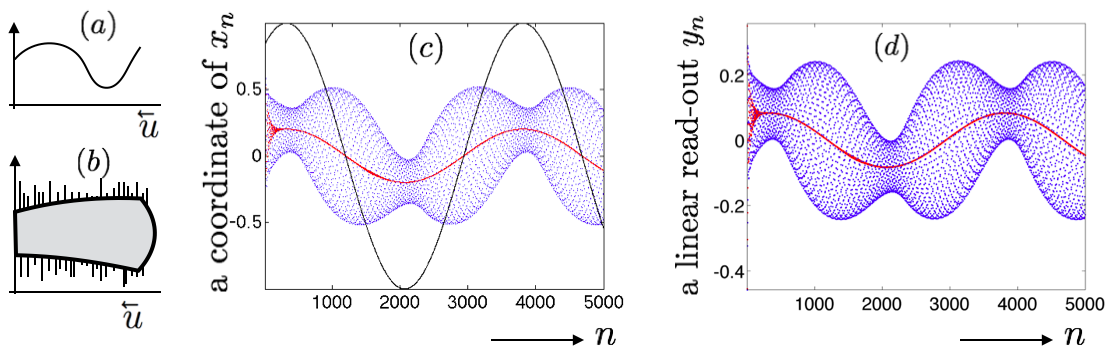}
\caption{Illustration of {\bf (R1)}:  (a) \& (b). Schematic of the graph of the input-encoding map when a system has the ESP in (a), and when it does not have the ESP in (b). The lighter grey values indicate the interior, while the darker values indicate the boundaries of the set-valued function's graph, and the comb-like outward projections refer to the explosions due to the absence of l.s.c.  (c). A coordinate of a simulated solution $(x_0,x_1,\ldots,x_{5000})$ plotted in red and blue against time $n$ (for an RNN
with a sinusoidal input shown in black) respectively when $\alpha=1.02$ (while it has ESP) and when $\alpha=1.05$ (while it does not have ESP); (d). A linear read-out $y_n$ of the states $x_n$ in (c) plotted against time $n$ in red and blue when $\alpha=1.02$ and when $\alpha=1.05$ respectively.} 
\label{Fig_2orbits}
\end{figure}
\end{center}

The scenarios when the system has ESP, and when it does not have the ESP w.r.t.\ an $\cev{u}$ is shown schematically in (a) and (b) of Fig.~\ref{Fig_2orbits}. At a point of discontinuity $\cev{u}$ of the input-encoding map, the map is u.s.c but not l.s.c at $\bar{u}$ which means that there is an explosion in the set $\mathcal{E}(\alpha, \cev{u})$ at $\bar{u}$ (see Lemma~\ref{Lemma_USC}). Thus the input-encoding map becomes very sensitive to a small change in input and behaves wildly as schematically indicated in (b) of Fig.~\ref{Fig_2orbits} (see \cite[Fig. S4]{SI} for another illustration). It is not feasible to simulate the input-representation map (or the input-encoding map) by fixing $\alpha$ and varying the inputs since each input lies in an infinite-dimensional space.  However, we demonstrate in (c) of Fig.~\ref{Fig_2orbits} 
that a sinusoidal input (in black) solely guides the dynamics of a driven system by plotting (in red) a coordinate of points when there is the ESP w.r.t.\ the input, and that the system response (a coordinate in blue) shows rapid fluctuations incommensurate with the input when it does not have the ESP w.r.t.\ the same sinusoidal input. During the training of RNNs in the RC framework, a linear observable of the solution that is designed is called a read-out. The read-out map is a linear transformation of the state $x_n$, and hence a continuous map of the state $x_n$. Whenever the input-encoding map is discontinuous, the collection of all possible read-outs as a function of the parameter $\alpha$ or the input $\bar{u}$, i.e., any such resultant read-out encoding map would then be a composition of a discontinuous map (the input-encoding or the parameter-encoding map) and a continuous map (coming from the linear read-out) map. We know that the composition of a discontinuous map and a continuous map is not continuous unless in very specific cases (like the read-out is a constant value).  The read-out is plotted against time in (d) of Fig.~\ref{Fig_2orbits}. We observe that when the ESP fails, the read-out also shows rapid fluctuations that are incommensurate with the input. Hence the chance of mitigating the effect of discontinuity of the input-encoding map or counter-balancing the distorted system response to input with a read-out is remote.

\begin{center}
\begin{figure}[h]
\centering
\includegraphics[width=15cm]{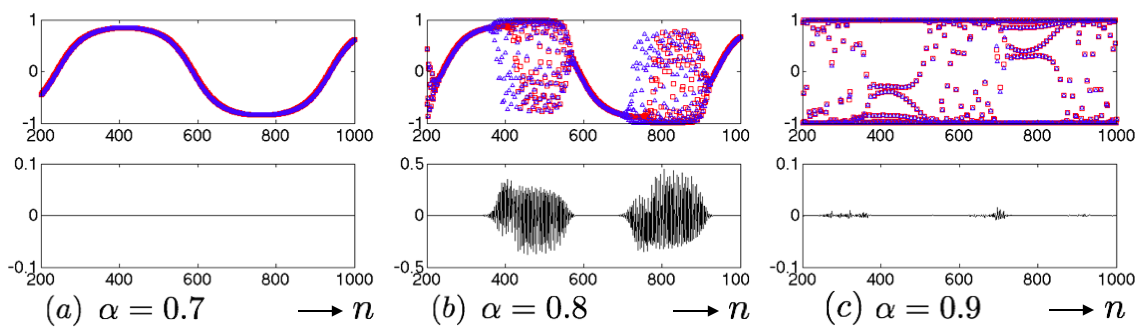}
\caption{Figure to illustrate nuances with input-related  stability: (Top Panel) A coordinate of a simulated solution $x_n$ of a RNN with a sinusoidal input $u_n$ (in blue) and a solution with the noisy input $v_n=u_n+\epsilon_n$ where $\epsilon_n$ is a realization of zero-mean uniformly distributed noise with maximum amplitude $10^{-3}$ while (a) $\alpha =0.7$, $\alpha=0.8$ and $\alpha=0.9$. (Bottom Panel) Difference between a linear read-out out of the solutions obtained from the inputs $u_n$ and $v_n$.}
\label{Fig_Quasi_Stability}
\end{figure}
\end{center}

We now illustrate {\bf (R1)} from the viewpoint of the sensitivity to input noise to indicate the importance of the ESP, and also to show that some networks may strangely exhibit some robustness to noise even without the ESP. We consider the matrix in \cite[Equation 6]{yildiz2012re} and scale it to have a unit spectral radius to use it as the matrix $B$ in a RNN ($B$ is as in \cite[Section 6]{SI}). It is known that there exist values of $\alpha<1$ so that
the RNN with such a $B$ does not have the ESP w.r.t.\ an input that is identically zero owing to a Neimark-Sacker bifurcation  \cite{yildiz2012re}.  
For the illustration considered, we remark that the system is found to have the ESP w.r.t.\ a sinusoidal input for $\alpha=0.7$ and not while $\alpha=0.8$, such an inference is drawn using a parameter-stability plot (discussed later). We show the response of the system for the sinusoidal input $u_n$ and an input $v_n=u_n+\epsilon_n$ where $\epsilon_n$ is a realization of uncorrelated noise uniformly distributed in the interval $[-10^{-3},10^{-3}]$. The responses (solutions) obtained from the same initial condition $x_0$ are shown in the top panel of Fig.~\ref{Fig_Quasi_Stability} along with the differences in the linear read-outs of the solutions in the bottom panel. It is remarkable that even with the noise of a magnitude not greater than $10^{-3}$, the differences in the solutions and hence the read-out is disproportionately large when the ESP fails at $\alpha=0.8$. When $\alpha=0.9$, the solutions do not have the sinusoidal shape; however, they are close to each other, and so are the read-outs. This quasi input-related stability depends on the network architecture and is observed to be exhibited only when the initial conditions $x_0$ (chosen to simulate the solutions) are from a proper subset of $X$. We believe that RNNs trained without the reservoir computing framework may exhibit such stability.  Such stability is not generic and dependent on the matrix $B$ and investigating into it is part of the author's ongoing research.

It may be noted that the continuity of the input-encoding map (or the input-representation map) is local because it is defined for every single input, and further, it can be defined regardless of whether the ESP w.r.t.\ the input is defined or not. The notion of the fading memory property \cite{boyd1985fading} is more of a global stability notion in the sense that it is defined on a subspace of input sequences for which the ESP w.r.t.\ all such inputs is a pre-requisite (see \cite{grigoryeva2018echo,grigoryeva2019differentiable}). Fading memory property and the input-related stability introduced here cannot be directly compared unless one extends the input-related stability to a global version, i.e., $g$ is said to have the global input-related stability at parameter $\alpha$ if the input-encoding map $\mathcal{E}(\alpha,\cdot)$ is continuous for all $\cev{u}$.  In that case, it is possible to show \cite{Manju_Prep} that the fading memory property and the global input-related stability are equivalent for an open-parametric driven system that is contractible. The same equivalence can also be extended to the input forgetting property.

Result {\bf (R2)} concerns parameter-related stability. It is clear from {\bf (R2)} that when parameter-related stability fails, i.e., when the parameter encoding map is discontinuous, then the driven system does not have the ESP w.r.t.\ the concerned input. Hence input-related stability fails as well by  {\bf (R1)}, and the  discussion presented in Fig.~\ref{Fig_2orbits} and Fig.~\ref{Fig_Quasi_Stability} holds when parameter-related stability fails.  A difference between parameter-related stability and input-related stability is that parameter-related stability is not necessarily equivalent to ESP as observed from {\bf (R2)}. In fact the converse statement of  {\bf (R2)} is not true (see examples in \cite[Section 4]{SI}). It can be observed that we have not used the contractibility hypothesis to conclude the result in {\bf (R2)}.

\begin{center}
\begin{figure}[h]
\centering
\includegraphics[width=12cm]{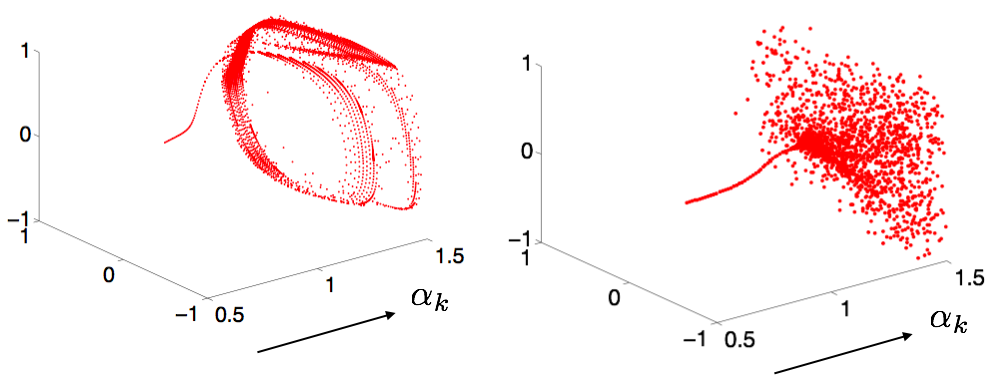}
\caption{A simulation of the parameter-encoding map: Given a RNN with multiple nodes and an input, we employ several initial conditions (samples) in $X$ to simulate elements in the set $\mathcal{E}(\cdot,\cev{u})$. Two coordinates the $\mathcal{E}(\cdot,\cev{u})$ are plotted against an increasing sequence $\alpha_k$. For the figure on the left, a reservoir with 2 neurons with 1000 samples in $X$ was used, and on the right a reservoir with 250 neurons and 50 samples in $X$ was used.}
\label{Fig_States_vs_Alpha}
\end{figure}
\end{center}

It is possible to simulate the parameter-encoding map when $\alpha$ is a scalar at discrete steps $\alpha_k$ to observe the scattering of system states while the ESP is not satisfied. 
When the ESP is satisfied w.r.t.\ $\cev{u}$, i.e., when  $\mathcal{E}(\cdot,\cev{u})$ is a singleton subset of $X$ for some $\alpha$, a numerical estimate of 
$\mathcal{E}(\cdot,\cev{u})$ is $\mathcal{E}_n(\alpha,\cev{u})$, and it would contain a single highly clustered set of points, and not so when  $\mathcal{E}(\cdot,\cev{u})$ is multi-valued i.e, when it is a subset of $X$ with multiple elements.
The result of identifying such a distinction can be made even with a small number of initial conditions. Fig.~\ref{Fig_States_vs_Alpha} shows two coordinates of the parameter-encoding map
$\mathcal{E}(\cdot,\cev{u})$ simulated at discrete set of points $\alpha_k$ for two different RNNs.
We consider $g(\alpha,u,x) = \overline{\tanh}(Au + \alpha Bx)$ with $2$ neurons and with $250$ neurons for the two RNNs, a randomly generated input of length $500$, a input matrix $A$ and a reservoir matrix $B$ (chosen randomly, but with unit spectral radius). For the plot on the left in 
Fig.~\ref{Fig_States_vs_Alpha}, we use 1000 initial conditions chosen on the boundary  of $X=[-1,1]^2$  (to exploit the fact that $g(\alpha,u,\cdot): X \to X$ is an open mapping \cite{munkres2014topology} and hence preserves the boundary). For the plot on the right, there were only $50$ chosen samples in $X=[-1,1]^{250}$.  Clearly, in both plots of Fig.~\ref{Fig_States_vs_Alpha}, the clustering towards a single-point for smaller values of $\alpha$ is obvious.  One does not need to be worried about simulating the parameter-encoding map accurately if one's task is only to identify if $\mathcal{E}(\cdot,\cev{u})$ has a single point-cluster.  Conversely, it is not possible to ascertain the parameter value at which the parameter-encoding map turns discontinuous or the parameter-related stability fails by using {\bf (R2)}. However, as we find in {\bf (R3)}, the continuity or discontinuity of the parameter-encoding map can be ascertained by the equicontinuous property of its finite time-approximations $\{\mathcal{E}_n(\cdot,\cev{u})\}$.

In the particular case, when $\alpha$ is a scalar parameter in an interval $[a,b]$, given an open-parametric driven system $g$ and an input $\cev{u}$, we define $\alpha_t \in [a,b]$ to be the ESP threshold if $g$ has the ESP w.r.t.\ $\cev{u}$ for all $\alpha < \alpha_t$ (or $\alpha > \alpha_t$) and does not have the ESP w.r.t.\ $\cev{u}$ for all $\alpha > \alpha_t$ (or $\alpha < \alpha_t$). Further, suppose $\alpha_t$ is the ESP threshold in an interval $[a,b]$, and the parameter-encoding map is also continuous at $\alpha_t$, we would call $\alpha_t$ to be a soft-ESP threshold, and call $\alpha_t$ a hard-ESP threshold when parameter-encoding map is discontinuous at $\alpha_t$ (examples are in \cite[Section 5]{SI}).  If $\alpha_t$ is a soft-ESP threshold, there would be a seamless transition of the parameter-encoding map from being single-valued  to being multi-valued (as in (a) of \cite[Fig. S3]{SI}) while $\alpha$ crosses the soft-ESP threshold and hence there would be no instabilities (discontinuous changes) in the system response due to fluctuations in $\alpha$.

The collective states of many interacting particles (elements) of a system show discontinuous behaviors when they undergo a ``phase transition" due to the variation of a parameter (e.g. \cite{kauffman1996home}). The parameters at which this happens are called critical points, and they mark a phase transition.  For the lack of a general mathematical description of a phase transition, a critical transition is often identified or defined by the accompanying changes during a physical transition. These changes could be such as the emergence of a power-law distribution of some entities involved or observing a long-range correlation between the particles or by observing a transition from ``order" to ``disorder". The issue with defining a critical transition based on the accompanying changes is that some of these changes can also be observed even while there is no criticality observed (e.g., \cite{beggs2012being}).  Here, within a mathematical framework, we are in a position to describe the collective states of a driven system at a parameter $\alpha$, especially if it also represents a network with many nodes, as the set $X_n(\alpha, \bar{u})$.  If we have an open-parametric driven system, then it is enough to understand a critical transition at $\alpha_0$ to be a sudden or discontinuous change in the set $\mathcal{E}(\alpha, \cev{u}) = X_0(\alpha, \bar{u})$ as $\alpha$ crosses $\alpha_0$.

Based on experiments focused on living systems, Kauffman \cite{kauffman1993origins,kauffman1996home} and on physical systems, Packard \cite{packard1988adaptation}, Langton et al. \cite{langton1990computation} and Crutchfield et al. \cite{crutchfield1988computation} conjectured that a dynamical regime between order and disorder optimally balances adaptiveness and robustness to input or stimuli.  Based on these works, we here make the edge-of-criticality to refect on a transition that shows a change in the robustness to the input noise.  By {\bf(R1)}, such a transition takes place exactly when the ESP is lost. Hence, we define the input-specific edge-of-criticality as the hard-ESP threshold, when both the parameter-encoding map (reflecting a change in the collective states) and the input-encoding map suddenly turn discontinuous.  Note that in addition to having made the notion of edge-of-criticality precise, we have made this notion input-specific as it should be since the parameter-encoding map is input-specific. Thus applying {\bf (R3)} to the case when $\alpha$ is a real-parameter in the interval $[a,b]$ with $\alpha_t$ as the ESP threshold, then $\alpha_t$  is a soft-ESP threshold when $\{\mathcal{E}_n(\cdot,\cev{u})\}$ is equicontinuous at $\beta$ and a hard- ESP threshold or the edge-of-criticality otherwise.

There are heuristic methods in the literature to approximate the edge-of-criticality owing to the lack of a clear definition of the edge-of-criticality. 
Result {\bf (R3)} or Theorem~\ref{Thm_Main2b} distinguishes the two ESP thresholds defined above and, more generally, the continuous or discontinuous points of the parameter-encoding map through the notion of equicontinuity.  Intuitively, equicontinuity is meant to deal with continuity of the
entire set of functions at once. Formally,
a family of functions $\{f_i\}_{i\in I}$ defined between two metric spaces $X$ and $Y$ is equicontinuous at $x\in X$ if for every $\epsilon>0$, there exists a $\delta >0$ such that $d_X(x,y) < \delta$ would imply 
$
d_Y(f_i(x),f_i(y)) < \epsilon$ for all $i\in I$.
The family $\{f_i\}_{i\in I}$ is said to be \emph{equicontinuous} on a set $A \subset X$ if it is equicontinuous at every point $x\in A$. A very simple example, is family of functions defined by $f_n(x)=x^n$, $n=1,2,\ldots$ on $[0,1]$ is not equicontinuous at $1$ since for instance when
$\epsilon=1/2$, for every $\delta>0$, there is an $n$ large enough so that $|f_n(y) - f_n(y)| = |1-y^n| > \epsilon$ while $y \in (1-\delta,1)$. The family of functions is equicontinuous for $x\in [0,1)$ though. If the parametric-encoding family $\{\mathcal{E}_n(\cdot, \cev{u})\}$ has a similar behavior, then we witness a discontinuity in the parameter-encoding map as Theorem~\ref{Thm_Main2b} asserts.  Owing to the property of non-equicontinuity associated with the hard-ESP threshold or the edge-of-criticality  
we can identify the threshold in a numerical simulation.

To illustrate this simulation, we simulate the parameter-encoding map at a discrete set of parameter values $(\alpha_1,\alpha_2,\ldots, \alpha_l)$ when $\alpha$ is a scalar in an interval $[a,b]$. We employ the definition of the Hausdorff metric $d_H$ (see Section~\ref{Sec_MD}). Given an input $\cev{u}$, an integer $n$,  we consider a plot of  $\gamma_n (\alpha_k) = d_H(\mathcal{E}_n(\alpha_{k},\cev{u}), \mathcal{E}_n(\alpha_{k-1},\cev{u}))$ against $\alpha_k$, which we call the parameter-stability plot and the procedure to obtain it is described next.

\begin{center}
\begin{figure}[h]
\centering
\includegraphics[width=14cm]{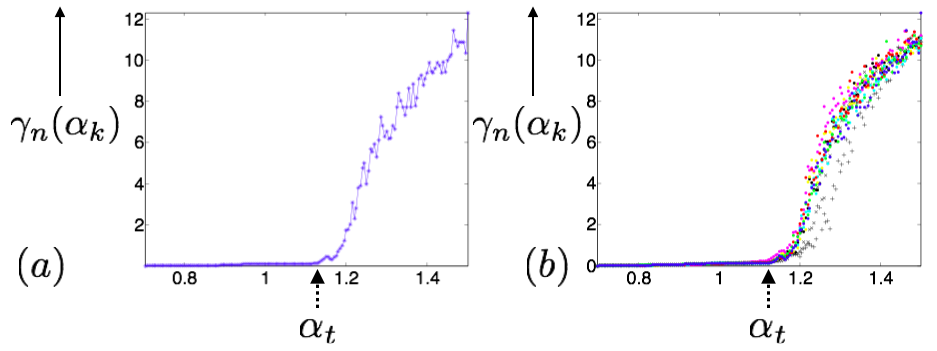}
\caption{Parameter-Stability plot(s) of an RNN: $\gamma_n (\alpha_k) = d_H(\mathcal{E}_n(\alpha_{k},\cev{u}), \mathcal{E}_n(\alpha_{k-1},\cev{u}))$ plotted against $\alpha_k$ with $|\alpha_{k}-\alpha_{k-1}| = 0.005$. (a) With a randomly generated $\cev{u}$ and $n=500$ (b) data sets of nine parameter-stability plots (each set with a different colour) obtained by the input $\sigma^j(\cev{u})$ where $j$ is varied from 0 to 80 in steps of 10, and with respective input lengths being varied as $n=500-j$. The smallest value of $\alpha_k$ where the plot turns distinctly positive is the estimate of the edge-of-criticality or the hard-ESP threshold $\alpha_t$ in the interval $[0.7,1.5]$.}
\label{Fig_ESP_Threshold}
\end{figure}
\end{center}

{\bf Procedure to obtain the parameter-stability plot.} A step-wise procedure to numerically identify the discontinuous points of the parameter-encoding map and the edge-of-criticality or hard-ESP threshold $\alpha_t$ for a given $\cev{u}$ (when the parameter is a scalar) is presented next (we employ this procedure to determine $\alpha_t$ in Fig.~\ref{Fig_ESP_Threshold}): \begin{enumerate}
\item Fix equally distant points $a=\alpha_1<\alpha_2<\cdots< \alpha_l=b$ in an interval $[a,b]$;
\item With $M$ initial conditions of the reservoir and considering $n$ successive values of the input, simulate the set $\mathcal{E}_n(\alpha_k,\cev{u})$. 
\item Find $\gamma_n(\alpha_k) = d_H(\mathcal{E}_n(\alpha_{k},\cev{u}), \mathcal{E}_n(\alpha_{k-1},\cev{u}))$; the Hausdorff distance $d_H(S_1,S_2)$ between two finite sets $S_1$ and $S_2$ with identical cardinality $M$ is computed practically by $\max(\max(\min(D)),\max(\min(D^T))$, where: (i) $\min(C)$ denotes the minimum of each row in a matrix $C$ (ii) the $(i,j)$ element of $D$ is 
the norm between the $i$th element of $S_1$ and $j$th element of $S_2$, i.e., 
$ 
D(i,j):=||S_1(i)-S_2(j)||_{p}$ (any $p$-norm could be employed in finite dimensions) (iii). $D^T$ is the transpose of $D$.
\item  Obtain the parameter-stability plot, i.e.,  $\gamma_n(\alpha_k)$ against $\alpha_k$ and identify the smallest $\alpha_j$ where the plot of $\gamma_n(\alpha_j)$ turns conspicuously positive as the edge-of-criticality or the hard-ESP threshold in the interval $[a,b]$, and other $\alpha_j$ where the plot is conspicuously positive as the points of discontinuities of the parameter-encoding map.  If the plot has only constant behavior close to $0$, then the parameter-encoding map is continuous everywhere. 
\end{enumerate}

While the parameter-encoding map is continuous at an $\alpha_k$, $\gamma_n(\alpha_k)$ can be made arbitrarily close to $0$ by setting $|\alpha_{k} - \alpha_{k-1}|$ appropriately, and while it is discontinuous at $\alpha_k$, $\gamma_n(\alpha(k))$ would turn positive  despite $|\alpha_{k} - \alpha_{k-1}|$  being small since the parameter-encoding family is not equicontinuous (by {\bf (R3)}) and due to the fact that  $\mathcal{E}_{n+1}(\alpha,\cev{u}) \subset \mathcal{E}_n(\alpha,\cev{u})$. Hence, we can identify the  smallest value of $\alpha_k$ at which the parameter-stability plot turns distinctly positive as an
estimate of the input-specific edge-of-criticality or the hard-ESP threshold. 

The parameter-stability plot of an RNN is presented in Fig.~\ref{Fig_ESP_Threshold} (a). In Fig.~\ref{Fig_ESP_Threshold} (b), we plot the data of nine parameter-stability plots that are obtained by $n$ successive values of the input, and different shifts of the input $j$, where the input and the RNN employed is the same as in 
(a) of Fig.~\ref{Fig_ESP_Threshold}.  By the definition of an open-parametric driven system, we know that $\alpha_t$ is an ESP threshold for a driven system with input $\cev{u}$ if and only if it is also the ESP threshold with the input $\sigma^j(\cev{u})$,  where the function $\sigma^j(\cdot)$ shifts each left-infinite sequence in its argument to the right by $j$ units.   We observe that $\alpha_t$ determined with the plot in Fig~\ref{Fig_ESP_Threshold} (b) is robust to both the different shifts and even shorter input lengths $n$  and corroborates with the estimate of $\alpha_t$ determined in  Fig~\ref{Fig_ESP_Threshold}. Further simulation results that show the robustness of the results due to variation in $|\alpha_{k} - \alpha_{k-1}|$ and the effect of input scaling are presented in \cite[Section 7]{SI}.

\section{Mathematical Details and Proofs} \label{Sec_MD}

We present the rigorous versions of the results {\bf (R1)}, {\bf (R2)} and {\bf (R3)} and prove them. The content here is also reproduced as three separate sections, along with additional examples in \cite[Sections 2,3,4]{SI}. In our analysis, wherever a cartesian product of spaces is considered, the cartesian products are endowed with the most commonly used topology called product topology (e.g., \cite{munkres2014topology}). We recall that a function taking values in a cartesian product equipped with product topology is continuous if and only if its coordinate functions are all continuous \cite{munkres2014topology}.

While we refer to a parametric driven system here, we bear in mind that it would comprise a parameter space $\Lambda$, an input space $U$, both topological spaces and a compact metric space $X$ along with a continuous map $g: \Lambda \times U \times X \to X$.  When $X$ is a
metric space with metric $d$, we denote by $\mathsf{H}_X$ the 
collection of all nonempty closed subsets of $X$. On the space $\mathsf{H}_X$ we equip the Hausdorff metric defined by
$d_H(A,B) :=  \inf\{ \epsilon : A
\subset B_\epsilon(B) \: \: \& \: \:B\subset B_\epsilon(A) \}$ where
$B_\epsilon(A): = \{ x \in X : d(x,A) < \epsilon\}$ is the open $\epsilon$-neighborhood of the set $A$. It is well known that $\mathsf{H}_X$ is compact when $X$ is compact.

The set $X_{n}(\alpha,\bar{u})$ defined previously through a solution of $g$ can be defined equivalently in forms that are more convenient to be employed in a proof.  Towards that end, we would adopt a composition-operator (called a process in \cite{Manju_ESP, kloeden2011nonautonomous}).  Given a parametric driven system $g$,  we denote $g_{\alpha,u}(x) := g(\alpha,u,x)$ and $\{u_n\} \subset U$ by $\bar{u}$. 
Suppose a parametric driven system $g$ has been fed input values $u_{m},u_{m+1},\ldots,u_{n-1}$ starting at time $m$. Then the map $g$ transports a state-value $x\in X$ at time $m$ to give a state-value $g_{\alpha,u_{n-1}}\circ \cdots \circ g_{\alpha,u_{m}}(x)$ at time $n$.  
Formally, for every choice of $\alpha$, $\bar{u}$ and $g$, we define for all $m\le n$, the function that `transports' a system state at $x$ at time $m$ through the inputs $u_{m}, u_{m+1}, \ldots, u_{n-1}$ to the state at time $n$ given by an composition-operator  $\phi_{\alpha,\bar{u}}: \mathbb{Z}^2_{\ge} \times X \to X$, where $\mathbb{Z}^2_{\ge} := \{ (n,m) : n \ge m, n,m \in \mathbb{Z} \}$ and
\begin{align*} 
  \phi_{\alpha,\bar{u}}(n,m,x) := \begin{cases}
        x \quad \quad \quad \quad \quad \quad \quad \quad \quad \text{ if $n=m$,}
        \\
        g_{\alpha,u_{n-1}}\circ \cdots \circ g_{\alpha,u_{m}}(x) \text{ if }  m< n.
        \end{cases}
\end{align*} 

Thus, if the inputs $u_{m},u_{m+1},\ldots,u_{n-1}$
are fed in that order to every $x \in X$, the system would have evolved at time $n$ to one of the states in $\phi_{\alpha,\bar{u}}(n,m,X)$. A simple observation is that the set inclusion $\phi_{\alpha,\bar{u}}(m+2,m,X) \subset \phi_{\alpha,\bar{u}}(m+2,m+1,X)$ holds for all $m\in \mathbb{Z}$ since  $g_{\alpha,u_{m+1}} \circ g_{\alpha,u_{m}}(X) \subset g_{\alpha,u_{m+1}}(X)$.  Based on this observation, it follows that if the entire left-infinite input $\{u_m\}_{m<n}$ would have been fed to every $x\in X$, then $\phi_{\alpha,\bar{u}}(n,m,X)$ is a decreasing sequence of sets, i.e., $\phi_{\alpha,\bar{u}}(n,m-1,X) \subset \phi_{\alpha,\bar{u}}(n,m,X)$ for any $m\le n$. Hence, if the entire left-infinite input $\{u_m\}_{m<n}$ had influenced the dynamics of the parametric driven system $g$, the system would have evolved at time $n$ to an intersection of a decreasing sequence of sets: 
\begin{equation} \label{eqn_Xn}
X_n(\alpha,\bar{u}) := \bigcap_{m<n} \phi_{\alpha,\bar{u}}(n,m,X).
\end{equation}
When $X$ is compact, each set $\phi_{\alpha,\bar{u}}(n,m,X)$ is a closed subset of $X$ and hence $X_n(\alpha,\bar{u})$ is a nonempty closed subset of $X$ (a proof is available in  \cite[Lemma 2.1]{Manju_ESP}). Note that $X_n(\alpha,\bar{u})$ is an element of the space $\mathsf{H}_X$.  Since $\mathsf{H}_X$ is a compact metric space, every sequence of sets has a convergent subsequence. Whenever a sequence $\{A_n\}$ converges in $\mathsf{H}_X$, it means that $d_H(A_n,A) \to 0$ and we denote it by $d_H\text{\,-}\lim_{n\to \infty} A_n = A$.  It may be verified that (see for e.g., \cite{aubin2009set}) when $\{A_n\}$ is a sequence of decreasing sets, i.e., $A_{n+1} \subset A_n$, then $A=\bigcap_{n=1}^\infty A_n$ if and only if $d_H\text{\,-}\lim_{n\to \infty} A_n = A$, i.e., the limit in the compact space 
$\mathsf{H}_X$ always exists for a decreasing sequence of sets and is equal to the nested intersection of the sequence of sets. Hence 
an alternate definition of \eqref{eqn_Xn} would be
\begin{equation} \label{eqn_Xn2}
X_n(\alpha,\bar{u}) := d_H\text{\,-}\lim_{m\to \infty}  \phi_{\alpha,\bar{u}}(n,-m,X).
\end{equation}

One can easily show  that ``a sequence $\{x_n\}$ is a solution of $g$ obtained through an input $\bar{u}$ if and only if $x_{k}$ belongs to $X_{k}(\alpha,\bar{u}) $ for all $k \in \mathbb{Z}$" (see \cite[Lemma 2.1]{Manju_ESP} for a proof).  Hence the definitions of $X_{n}(\alpha,\bar{u})$ that we have considered are all equivalent.

The continuity of the function $(\alpha,\bar{u}) \mapsto X_{n}(\alpha,\bar{u})$ can be defined by considering the topology induced on $\mathsf{H}_X$ by the Hausdorff metric $d_H$. Such continuity defined using the Hausdorff metric is equivalent \cite{aubin2009set} to the continuity of 
$X_{n}(\alpha,\bar{u})$ when $X_{n}(\alpha,\bar{u})$  is treated as a set-valued function of the variable $(\alpha,\bar{u})$. We make use of this equivalence without further remarks. To define the continuity of a set-valued map, let $X$ be a topological space and $Y_X$ be a subspace of the power set of $X$, and let $Z$ be another topological space. A set-valued function $S: Z \to Y_X$  is continuous \cite{aubin2009set} at a point $x$ if it is both upper-semicontinuous and lower-semi-continuous at $x$. A function $S$ is upper-semicontinuous (u.s.c) at $x$ if for every open set $V$ containing $S(x)$ there is a neighborhood $U$ of $x$ so that $S(U)$ is contained in $V$; $S$ is lower-semicontinuous at $x$ (l.s.c) if for every open set $V$ that intersects $S(x)$ there is a neighborhood $U$ of $x$ so that $S(x)$ also has non-empty intersection with $V$ for all $x\in U$. (see \cite[Fig. S3]{SI} for a schematic picture).  If $S$ is u.s.c but not l.s.c at $x$ then it has an ``explosion" at $x$, and if  $S$ is l.s.c but not u.s.c at $x$ then it has an ``implosion" at $x$. The function $S$ is called continuous or u.s.c or l.s.c if it is respectively continuous, u.s.c and l.s.c at all points $x$ in its domain.  

We use the fact that if $S$ is both u.s.c at $x$ and is single-valued, it is continuous at $x$ \cite{aubin2009set}. An example of a map that is u.s.c but not l.s.c is the input-encoding map of a driven system when it does not have the ESP (see Theorem~\ref{Thm_Main1} and (ii) of Lemma~\ref{Lemma_USC}). We point out that the ``graphs" of such set-valued maps could behave wildly. It is not possible to explicitly describe such maps where the arguments are infinite-dimensional. An accessible example with a scalar argument is the set-valued function $f$ defined by
\begin{align} \label{eqn_usc}
  f(x) := \begin{cases}
        \{x\} \quad \quad \quad \quad \quad \quad \quad \text{ if $x$ is irrational or $x=0$,}
        \\
        [-\frac{1}{n} \,,\, \frac{1}{n}] \quad \quad  \text{ if }  x=m/n, m\in \mathbb{Z}, n \in \mathbb{N} \text{ are coprime.}
        \end{cases}
\end{align}
The function $f$ is u.s.c for all $x$ but l.s.c only when $x$ is irrational and at $x=0$. 

Given a parametric dynamical system $g$, for every given $\alpha \in \Lambda$, we call the set-valued mapping $\bar{u} \mapsto \{X_n(\alpha,\bar{u})\}_{n\in \mathbb{Z}}$ to be the \emph{input-representation map}.  The following theorem concerns input-related stability (version of {\bf (R1)}).

\begin{Theorem}{{\bf (Continuity of the input-representation map and the input-encoding map)}} \label{Thm_Main1} An open-parametric driven system $g$ that is contractible at $\alpha$  has the ESP w.r.t.\ $\bar{u}$ at $\alpha$ $\Longleftrightarrow$ the input-representation map  $\bar{u} \mapsto \{X_n(\alpha,\bar{u})\}$  is continuous at $\bar{u}$ $\Longleftrightarrow$ the input-encoding map
 $\bar{u} \mapsto \mathcal{E}(\alpha, \cev{u})$ is continuous at $\bar{u}$.
\end{Theorem}

The proof of Theorem~\ref{Thm_Main1} would be presented after establishing the relationship of the ESP w.r.t.\ $\bar{u}$ and the input-encoding map as stated in Lemma~\ref{Lemma_Main}. The pedantic reader may note that Theorem~\ref{Thm_Main1} would not be true without the contractiblility condition. For instance, we recall an example from Section~\ref{Sec_Results} that if $X=[0,1]$ and $g(\alpha,u,x) = x$, for all $n$, $X_n(\alpha, \bar{u})=X$ regardless of the input $\bar{u}$ and hence the input-representation map   $\bar{u} \mapsto \{X_n(\alpha,\bar{u})\}$ is continuous (i.e., both u.s.c and l.s.c), but there is no ESP to all inputs since  $X_n(\alpha, \bar{u})$ is not a singleton subset of $X$ for every $n$.

\begin{Lemma}{{\bf (Continuity of the input-encoding map)}} \label{Lemma_Main} An open-parametric-driven system $g$ that is contractible at $\alpha$  has the ESP w.r.t.\ $\cev{u}$ at $\alpha$ if and only if the input-encoding map $\mathcal{E}(\alpha,\cdot)$ is continuous at $\cev{u}$. 
\end{Lemma}

{\bf Proof. } ($\Longrightarrow$) The open-parametric driven system $g$ has the ESP w.r.t.\ $\cev{u}$ at $\alpha$ is equivalent to saying that  $\mathcal{E}(\alpha,\cev{u})$ is a singleton subset of $X$. By (ii) of Lemma~\ref{Lemma_USC}, $\mathcal{E}(\alpha,\cdot)$ is u.s.c at $\cev{u}$.   We use the fact that a map that is u.s.c and is a singleton subset of the space $X$ is continuous to conclude $\mathcal{E}(\alpha,\cdot)$ is a continuous at $\cev{u}$.  

($\Longleftarrow$) Let $\mathcal{E}(\alpha,\cdot)$ be continuous at $\cev{u}$ and suppose that $g$ does not have the ESP w.r.t.\ $\cev{u}$ at $\alpha$. This means  $\mathcal{E}(\alpha,\cev{u})$ contains a subset of $X$ that has at least two elements of $X$.   Hence the encoding map evaluated at $(\alpha,\cev{u})$ is bounded away from all singleton subsets of $X$, i.e., $r := \inf_{\{x\} \in X} d_H(\mathcal{E}(\alpha,\cev{u}), \{x\}) > 0$.  

Since $g$ is contractible at $\alpha$, let $\cev{v}$ be an input for which $\mathcal{E}(\alpha,\cev{v})$ is a singleton subset of $X$. We define the sequence of left-infinite sequences $\{\cev{w}_n\}_{n > 1}$ by  
$\cev{w}_n : = (\ldots, v_{-2},v_{-1},u_{-{n}},\ldots,u_{-2},u_{-1})$. 
Since, the first $n$ elements of $\cev{w}_n$ are identical to that of $\cev{u}$, it follows that the sequence $\cev{w}_n \to \cev{u}$ in the product topology on $U^{(-\infty,-1)}$, where  $U^{(-\infty,-1)}$ is the infinite cartesian product space $\prod_{i=-1}^\infty U_i$ with $U_i=U$.

Since $\mathcal{E}(\alpha,\cev{v})$ is a singleton subset of $X$, we find that $\mathcal{E}(\alpha,\cev{w}_n)$ is a singleton subset of $X$ since
$$
\mathcal{E}(\alpha,\cev{w}_n) = \bigcap_{j>0} \phi_{\alpha, *_n}(0,-j,X) = \bigcap_{j>0} \phi_{\alpha, \dagger}(0,-j,X) = \mathcal{E}(\alpha,\cev{v}),
$$
where $*_n = \cev{w}_n$ and $\dagger = \cev{v}$ (notations to avoid obscurity in the subscripts).
Since $\mathcal{E}(\alpha,\cev{w}_n)$ is a singleton subset of $X$, it follows that $d_H(\mathcal{E}(\alpha,\cev{u}),\mathcal{E}(\alpha,\cev{w}_n)) \ge r> 0$ for all $n$. This implies $\mathcal{E}_n(\alpha,\cev{w}_n)$ does not converge to $\mathcal{E}(\alpha,\cev{u})$ although we had  $\cev{w}_n \to \cev{u}$. This contradicts our assumption that $\mathcal{E}(\alpha,\cev{u})$ is continuous at $\cev{u}$. 
$\blacksquare$.

\begin{Corollary} \label{Corollary_Continuity}
If an open-parametric-driven system $g$ that is not necessarily contractible at $\alpha$ has the ESP w.r.t.\ $\cev{u}$ at $\alpha$ then the input-encoding map $\mathcal{E}(\alpha,\cdot)$ is continuous at $\cev{u}$. 
\end{Corollary}

{\bf Proof of Corollary~\ref{Corollary_Continuity}. } The first paragraph of the proof of Lemma~\ref{Lemma_Main}.

{\bf Proof of Theorem~\ref{Thm_Main1}. } From Lemma~\ref{Lemma_Main}, we have: an open-parametric driven system $g$ that is contractible at $\alpha$  has the ESP w.r.t.\ $\bar{u}$ at $\alpha$ if and only if  the input-encoding map
 $\bar{u} \mapsto \mathcal{E}(\alpha, \cev{u})$ is continuous at $\bar{u}$. It remains to be proven that the input-encoding map
 $\bar{u} \mapsto \mathcal{E}(\alpha, \cev{u})$ is continuous at $\bar{u}$ if and only if the input-representation map  $\bar{u} \mapsto \{X_n(\alpha,\bar{u})\}$  is continuous at $\bar{u}$. Equip the representation space $R:=\prod_{i=-\infty}^\infty Y_i$, where $Y_i=  \mathsf{H}_X$ for all $i\in \mathbb{Z}$ with the product topology. Let $\cev{v}_n := (\ldots, u_{n-2},u_{n-1})$. Clearly, 
$X_n(\alpha, \bar{u}) = \mathcal{E}(\alpha, \cev{v}_n)$ and the mapping $\bar{u} \mapsto X_n(\alpha,\bar{u})$ is identical to $\cev{v}_n \to \mathcal{E}(\alpha, \cev{v}_n)$. The following equivalent statements (i)--(vi) lead to the proof:
(i) $\bar{u} \mapsto \{X_n(\alpha,\bar{u})\}$ is continuous if and only if (iff) the coordinate mappings $\bar{u} \mapsto X_n(\alpha,\bar{u})$ is continuous for all $n\in \mathbb{Z}$ (by definition of convergence in product topology) (ii).
$\bar{u} \mapsto X_n(\alpha,\bar{u})$ is continuous for all $n\in \mathbb{Z}$ iff 
$\cev{v}_n \mapsto \mathcal{E}(\alpha, \cev{v}_n)$ is continuous for all $n\in \mathbb{Z}$ (by definition of $\cev{v}_n$). (iii). $\cev{v}_n \mapsto \mathcal{E}(\alpha, \cev{v}_n)$ is continuous for all $n\in \mathbb{Z}$ iff  $g$ has the ESP w.r.t.\ $\cev{v}_n$ for all $n \in \mathbb{Z}$ (by Lemma~\ref{Lemma_Main}). (iv). $g$ has the ESP w.r.t.\ $\cev{v}_n$ for all $n \in \mathbb{Z}$ iff 
$g$ has the ESP w.r.t.\ $\cev{v}_0$  (since $g$ is open-parametric driven system) (v). $g$ has the ESP w.r.t.\ $\cev{v}_0$ iff   $g$ has the ESP w.r.t.\ $\cev{u}$ (by definition of $\cev{v}_n$). (vi.)  $g$ has the ESP w.r.t.\ $\cev{u}$ iff the encoding map $\cev{u} \mapsto \mathcal{E}(\alpha,\cev{u})$ is continuous (by Lemma 1). $\blacksquare$

We remark that the topology on representation space $R$ can be obtained through a metric $d_R$ \cite{munkres2014topology}. An example is  $d_R(A,B) := \sum_{i=-\infty}^{\infty} d_H(A,B)/2^{|i|}$. An analogously defined metric $d_U$ can be used as a metric for the input sequence space $U^{(-\infty,+\infty)}$. We make use of  Lemma~\label{Lemma_USC} that follows from the assumption that $g$ is continuous and $X$ is compact.  Lemma~\label{Lemma_USC} is proved in \cite[Lemma 2]{SI}.

\begin{Lemma} \label{Lemma_USC}
Let $g$ be a driven dynamical system.  Then
\renewcommand{\labelenumi}{(\roman{enumi}).}
\begin{enumerate}
\item The parameter-encoding map $\mathcal{E}(\cdot, \cev{u})$ is u.s.c for a given input $\cev{u}$.

\item The input-encoding map $\mathcal{E}(\alpha, \cdot)$ is u.s.c for a given $\alpha$.

\item Every function in the input-encoding family 
$\{\mathcal{E}_n(\cdot, \cev{u})\}$ or in  the parameter-encoding family $\{\mathcal{E}_n(\cdot, \cev{u})\}$ is continuous.

\end{enumerate}
\end{Lemma}

The following theorem establishes that the ESP ensures the continuity of the parameter-encoding map and is a rigorous version of {\bf (R2)}.

\begin{Theorem}{{\bf (Continuity of the parameter-encoding and the parameter-representation map)}} \label{Thm_Main2} 
Fix an $\bar{u}$ and hence $\cev{u}$. If an open-parametric driven system $g$  has the ESP w.r.t.\ $\cev{u}$ at $\alpha$ then the 
parameter-encoding map $\mathcal{E}(\cdot,\cev{u})$ is continuous at $\alpha$. More generally, if 
$g$  has the ESP w.r.t.\ $\bar{u}$ at $\alpha$   
then the parameter-representation map  $\alpha \mapsto \{X_n(\alpha,\bar{u})\}$ is continuous at $\alpha$.
\end{Theorem}

{\bf Proof. } Fix an $\bar{u}$ and hence the corresponding $\cev{u}$.  We note that from (i) of Lemma~\ref{Lemma_USC}, a parameter-encoding map is u.s.c. Further, whenever it is single-valued it is continuous. Hence $\mathcal{E}(\cdot,\cev{u})$ is continuous at $\alpha$.

Note that 
$\{X_n(\alpha,\bar{u})\} \in R := \prod_{i=-\infty}^\infty Y_i$, where $Y_i=  \mathsf{H}_X$ for all $i$ and (the representation space) $R$ is equipped with the product topology.
We know that $\alpha \mapsto \{X_n(\alpha,\bar{u})\}$ is continuous if and only if the coordinate mappings $\alpha \mapsto X_n(\alpha,\bar{u})$ is continuous for all $n\in \mathbb{Z}$. Let $\cev{v}_n := (\ldots, u_{n-2},u_{n-1})$. Clearly, 
$X_n(\alpha, \bar{u}) = \mathcal{E}(\alpha, \cev{v}_n)$ and the mapping $\alpha \mapsto X_n(\alpha,\bar{u})$ is identical to the parameter-encoding map $\alpha \to \mathcal{E}(\alpha, \cev{v}_n)$.  From Lemma~\ref{Lemma_USC}, a parameter-encoding map is always u.s.c, and when it is single-valued it is continuous. Since, $g$ is an open-parametric dynamical 
system, $\mathcal{E}(\alpha, \cev{v}_n)$ is a singleton subset of $X$ for all $n$ if and only if $\mathcal{E}(\alpha, \cev{u})$ is a singleton subset of $X$. When $g$ has the ESP w.r.t.\ $\cev{u}$ at $\alpha$ it follows that $\alpha \mapsto \{X_n(\alpha,\bar{u})\}$ is continuous. $\blacksquare$

We recall from Section~\ref{Sec_Results} that when we restrict the parameter space to be an interval $[a,b]$ the input-specific soft-ESP threshold and hard-ESP threshold as a parameter value $\beta \in [a,b]$ for which the parameter-encoding map is continuous and discontinuous respectively in the scenario where $g$ has the ESP w.r.t.\ $\cev{u}$ for all parameter values either to the left or right of $\beta$ (examples are in \cite[Section 4]{SI}). The following theorem establishes conditions on $\{\mathcal{E}_n(\cdot,\cev{u})\}$ and is a rigorous version of the result {\bf (R3) }.

\begin{Theorem} \label{Thm_Main2b} Consider an open-parametric-driven system $g$ and an input $\cev{u}$.  
The parameter-encoding map $\mathcal{E}(\cdot,\cev{u})$ is continuous at $\beta$ if and only if $\{\mathcal{E}_n(\cdot,\cev{u})\}$ is equicontinuous at $\beta$.  In particular, when $\alpha$ is a real-parameter in the interval $[a,b]$ and $\alpha_t$ is the ESP threshold, then $\alpha_t$  is a soft-ESP threshold when $\{\mathcal{E}_n(\cdot,\cev{u})\}$ is equicontinuous at $\beta$ and a hard-threshold otherwise.
\end{Theorem}

{\bf Proof. } We first show that $\{\mathcal{E}_n(\cdot,\cev{u})\}$ is equicontinuous at $\beta$ implies $\mathcal{E}(\cdot,\cev{u})$ is continuous at $\beta$.  To prove this it is sufficient to show that given a sequence,
$\alpha_k \to \beta$ we have $\mathcal{E}(\alpha_k, \cev{u}) \to \mathcal{E}(\beta, \cev{u})$. Fix a sequence $\alpha_k \to \beta$.  We know by definition of the sets
$\mathcal{E}_n(\alpha,\cev{u})$, $\mathcal{E}_n(\alpha,\cev{u}) \to \mathcal{E}(\alpha,\cev{u})$ for every $\alpha$. Hence for a given $\epsilon>0$, we can define $N(\alpha)$ to be an integer such that $d_H(\mathcal{E}(\alpha, \cev{u}), \mathcal{E}_n(\alpha, \cev{u})) < \epsilon/3$ for all $n>N(\alpha)$. 
Thus for every $k \in \mathbb{N}$, we deduce
\begin{eqnarray*}
d_H(\mathcal{E}(\alpha_k, \cev{u}), \mathcal{E}(\beta, \cev{u})) & \le & d_H(\mathcal{E}(\alpha_k, \cev{u}), \mathcal{E}_n(\alpha_k,\cev{u})) + d_H(\mathcal{E}_n(\alpha_k,\cev{u}), \mathcal{E}_n(\beta, \cev{u})) + \\
& & \quad \quad \quad  \quad \quad \quad \quad \quad \quad \quad \quad \quad  d_H(\mathcal{E}_n(\beta,\cev{u}), \mathcal{E}(\beta, \cev{u})) \\
& \le & \quad   \epsilon/3 \quad  + \quad  d_H(\mathcal{E}_n(\alpha_k,\cev{u}), \mathcal{E}_n(\beta, \cev{u})) \quad  +  \quad \epsilon/3,
\end{eqnarray*}
whenever $n>max(N(\alpha_k), N(\beta))$. 
Since $\{\mathcal{E}_n(\cdot, \cev{u})\}$ is equicontinuous at $\beta$, there exists an integer $K$ such that for $k>K$, $d_H(\mathcal{E}_n(\alpha_k,\cev{u}), \mathcal{E}_n(\beta, \cev{u})) \le \epsilon/3$. Hence, $\mathcal{E}(\alpha_k, \cev{u}) \to \mathcal{E}(\beta, \cev{u})$.

Next, to show $\mathcal{E}(\cdot,\cev{u})$ is continuous at $\beta$ implies $\{\mathcal{E}_n(\cdot,\cev{u})\}$ is equicontinuous at $\beta$, we prove its contrapositive. Let $\{\mathcal{E}_n(\cdot,\cev{u})\}$ be not equicontinuous at some fixed $\beta$. Then, there exists an $\epsilon>0$, a sequence $\alpha_k \to \beta$ and a sequence of integers $n_k \to \infty$ such that 
$d_H(\mathcal{E}_{n_{k}}(\alpha_k,\cev{u}), \mathcal{E}_{n_{k}}(\beta, \cev{u})) > 3\epsilon_0>0$. Hence, by triangle inequality,
\begin{eqnarray}
3\epsilon_0 \: < \: d_H(\mathcal{E}_{n_{k}}(\alpha_k,\cev{u}), \mathcal{E}_{n_{k}}(\beta, \cev{u})) & \le & d_H(\mathcal{E}_{n_{k}}(\alpha_k, \cev{u}), \mathcal{E}(\alpha_k,\cev{u})) \: \quad +  \label{eqn_noneq}\\
& & 
 d_H(\mathcal{E}(\alpha_k, \cev{u}), \mathcal{E}(\beta, \cev{u})) + d_H(\mathcal{E}(\beta, \cev{u}), \mathcal{E}_{n_{k}}(\beta, \cev{u})). \nonumber 
\end{eqnarray}

Since $\mathcal{E}_n(\beta, \cev{u}) \to \mathcal{E}(\beta, \cev{u})$, given $\epsilon_0 > 0$, we can find an integer $M$ such that the inequality  $d_H(\mathcal{E}_M(\beta, \cev{u}),\mathcal{E}(\beta, \cev{u})) < \epsilon_0$ holds. 
Note $\mathcal{E}_M(\cdot, \cev{u})$ is continuous for any finite $M$ by (iii) of Lemma~\ref{Lemma_USC}. Assume $\mathcal{E}(\cdot, \cev{u}))$ is continuous at $\beta$. Hence, by continuity
the set $V_\beta:= \{\alpha \in \Lambda : 
d_H(\mathcal{E}_M(\alpha, \cev{u}),\mathcal{E}(\alpha, \cev{u})) < \epsilon_0 \}$ contains a neighborhood of $\beta$ in the space $\Lambda$. Since 
$\mathcal{E}_n(\beta, \cev{u})$ is  monotonically decreasing, $d_H(\mathcal{E}_n(\beta, \cev{u}),\mathcal{E}(\beta, \cev{u})) \to 0$ monotonically. Hence $V_\beta:= \{\alpha \in \Lambda : 
d_H(\mathcal{E}_n(\alpha, \cev{u}),\mathcal{E}(\alpha, \cev{u})) < \epsilon_0 \: \forall \: n\ge M\}$. There exists an integer $K_1$ so that for all $k \ge K$, we have $\alpha_k \in V_\beta$ and $n_k\ge M$. Hence  the term $d_H(\mathcal{E}_{n_{k}}(\alpha_k, \cev{u}), \mathcal{E}(\alpha_k,\cev{u}))$ in \eqref{eqn_noneq} can be made less than $\epsilon_0$ whenever $k\ge K_1$. Thus, for $k\ge K_1$ we can rewrite 
\eqref{eqn_noneq} as
 \begin{eqnarray*}
3\epsilon_0 \: < \: d_H(\mathcal{E}_{n_{k}}(\alpha_k,\cev{u}), \mathcal{E}_{n_{k}}(\beta, \cev{u})) & \le & \epsilon_0 \: \quad + \\
& & 
 d_H(\mathcal{E}(\alpha_k, \cev{u}), \mathcal{E}(\beta, \cev{u})) + d_H(\mathcal{E}(\beta, \cev{u}), \mathcal{E}_{n_{k}}(\beta, \cev{u})). \nonumber 
\end{eqnarray*}
Now, by the assumption of continuity of $\mathcal{E}(\cdot, \cev{u})$ at $\beta$, there exists an integer  $K_2$ so that 
$d_H(\mathcal{E}(\alpha_k, \cev{u}), \mathcal{E}(\beta, \cev{u})) < \epsilon_0$. Also since $\mathcal{E}_{n_{k}}(\beta, \cev{u})) \to \mathcal{E}(\beta, \cev{u})$, there exists an integer $K_3$ so that $d_H(\mathcal{E}(\beta, \cev{u}), \mathcal{E}_{n_{k}}(\beta, \cev{u})) < \epsilon_0$. For $k \ge \max(K_1,K_2,K_3)$, we have $3\epsilon_0 < \epsilon_0+ \epsilon_0 + \epsilon_0$ which is absurd. Hence our assumption  $\mathcal{E}(\cdot, \cev{u}))$ is continuous at $\beta$ is incorrect. When $\Lambda =[a,b]$ and $\alpha_t$ is the ESP threshold, setting $\beta = \alpha_t$ in the above proof, we obtain $\alpha_t$ to be the soft-ESP threshold and the hard-ESP threshold
 when  $\{\mathcal{E}_n(\cdot,\cev{u})\}$ is equicontinuous and not-equicontinuous respectively at $\alpha_t$. $\blacksquare$

\section{Applications, Significance and Discussion} \label{Sec_Discussion} 

With computation, dynamics, and information being intermingled concepts, we find some interconnections between them.  The \emph{manifold hypothesis} (e.g., \cite{geron2019hands}) is based on the view that phenomena in the physical world more often lie in or around a lower-dimensional manifold that can be embedded in a higher-dimensional manifold. This has been observed in several fields, and in particular, well-explored in machine learning \cite{geron2019hands}.   When it comes to driven dynamics as considered in this article, the input $\bar{u}$ renders the system compute a representation $\{X_n(\alpha,\bar{u})\}$ that is usually a low-dimensional manifold, attracting (see pullback attractors in \cite{kloeden2011nonautonomous}) and lies in the higher-dimensional manifold $X$. In a broad class of driven systems, we establish in {\bf (R1)} that a continuous deformation of such a low-dimensional manifold (such manifolds exist, e.g., \cite{hart2019embedding}) is feasible by feeding close-by variants of the input if and only if the system has the ESP w.r.t.\ to the input that aided the computation of the low-dimensional manifold. Such continuous deformations are necessary for stable computations in all systems that compute through driven dynamics. From the perspective of systems that are controlled or driven, result {\bf (R1)} 
explains why memory-loss is needed to avoid unexpected behavior due to small changes in the temporal input.

Further, if parameters are a part of the design of driven systems, the representations are continuously deformed by a change in the parameter when the ESP is satisfied (result {\bf (R2)}). Thus in a neurocomputing scheme, if the interconnections of the network are set as a design-parameter, then the ESP ensures stability due to small changes in interconnection strengths in the network.  This result explains the empirically observed robustness of echo state networks with regard to the random choice of reservoir or input matrices in applications. We remark that such stability was explained in the context of a class of standard feedforward neural networks \cite{huang2006extreme}, but no such result is currently available for RNNs -- an anonymous referee has pointed out interesting error estimates recently obtained for RNNs in \cite{gonon2020approximation}.  ESP ensuring parameter-stability helps also design or explain stability in fully trained RNNs \cite{giles1994pruning}, and newer adaptations like deep recurrent neural networks \cite{hermans2013training}, conceptors \cite{jaeger2017using}, data-based evolving network models \cite{Manjunath_NC2} and in nonlinear open-loop control with a wide range of applications \cite{terrell2009stability}.

In the context, where parameters need to be pushed to the \emph{limit} to have better separability of inputs in the state space, it is important to identify this limit as the edge-of-criticality or the hard-ESP threshold. Unlike all earlier literature that describes the edge-of-criticality with phrases such as the boundary between order and chaos, or the midpoint between death and epilepsy, we define the input-specific edge-of-criticality mathematically owing to which we employ a procedure for numerically determining it using  {\bf (R3)}.  In the RC literature on RNNs, upper bounds on the attributes of the reservoir matrix (like spectral radius) to ensure the ESP for all possible inputs, i.e., the ESP w.r.t.\ to the entire space are available (e.g., \cite{yildiz2012re}). 
More commonly, the criterion of the unit spectral radius of the matrix $\alpha B$ is currently being used to have the ESP  w.r.t.\ all possible inputs.  
The disadvantage of these bounds or criteria is that they are not dependent on the input and particularly on the input's temporal properties.   In a practical application, often, only a class of inputs with specific templates are employed for a given task for the network. Hence the bounds are suboptimal on two accounts: (i) temporal relation in the input is ignored (ii) the gap between say the actual edge-of-criticality and the bound (like the unity of spectral radius of the reservoir) is unknown and we overcome this by using the parameter-stability plot. We anticipate that our procedure to determine the edge-of-criticality can be extended to a vector parameter in which case we could get a curve, surface, or a hypersurface as the hard-ESP threshold boundary.  From the viewpoint of stochastic inputs, we remark that when inputs are drawn from a stationary ergodic process,  the edge-of-criticality inferred from a single typical input is valid for all other generic inputs since the ESP holds with only either probability $0$ or $1$ \cite{Manju_ESP}. This would mean when the hard-ESP threshold is obtained from a typical realization of an ergodic process as the input, the same threshold holds for other realizations with probability 1.


\bibliographystyle{pnas-new}
\bibliography{pnas}

\end{document}